\documentclass{article} 
\usepackage{iclr2020_conference,times}

\usepackage{amsmath}
\usepackage{amssymb}
\usepackage{booktabs}
\usepackage{graphicx}

\usepackage{hyperref}
\usepackage{url}

\usepackage{pgfplots, pgfplotstable}
\usepackage{footnote}
\usepackage{floatrow}
\makesavenoteenv{tabular}
\makesavenoteenv{table}
\newfloatcommand{capbtabbox}{table}[][\FBwidth]

\usepackage{blindtext}

\DeclareMathOperator*{\argmax}{argmax}
\DeclareMathOperator*{\neighbor}{neighbor}
\DeclareMathOperator*{\simfunc}{sim}

\title{Multilingual Alignment of Contextual Word Representations}

\author{Steven Cao, Nikita Kitaev \& Dan Klein \\
	Computer Science Division \\
	University of California, Berkeley \\
	\texttt{\{stevencao,kitaev,klein\}@berkeley.edu}
}

\pgfplotstableread{
	X Y
	19.5 68.7
	26.1 70.4
	13.9 67.0
	32.5 74.5
	28.3 73.4
}\datatable
\iclrfinalcopy 
\begin{document}

	\maketitle
	
	\begin{abstract}
		
		We propose procedures for evaluating and strengthening contextual embedding alignment and show that they are useful in analyzing and improving multilingual BERT. In particular, after our proposed alignment procedure, BERT exhibits significantly improved zero-shot performance on XNLI compared to the base model, remarkably matching pseudo-fully-supervised translate-train models for Bulgarian and Greek. Further, to measure the degree of alignment, we introduce a contextual version of word retrieval and show that it correlates well with downstream zero-shot transfer. Using this word retrieval task, we also analyze BERT and find that it exhibits systematic deficiencies, e.g.\ worse alignment for open-class parts-of-speech and word pairs written in different scripts, that are corrected by the alignment procedure. These results support contextual alignment as a useful concept for understanding large multilingual pre-trained models.
	\end{abstract}
	
	\section{Introduction}
	\begin{figure}[h]
		\centering
		\includegraphics[width=\linewidth]{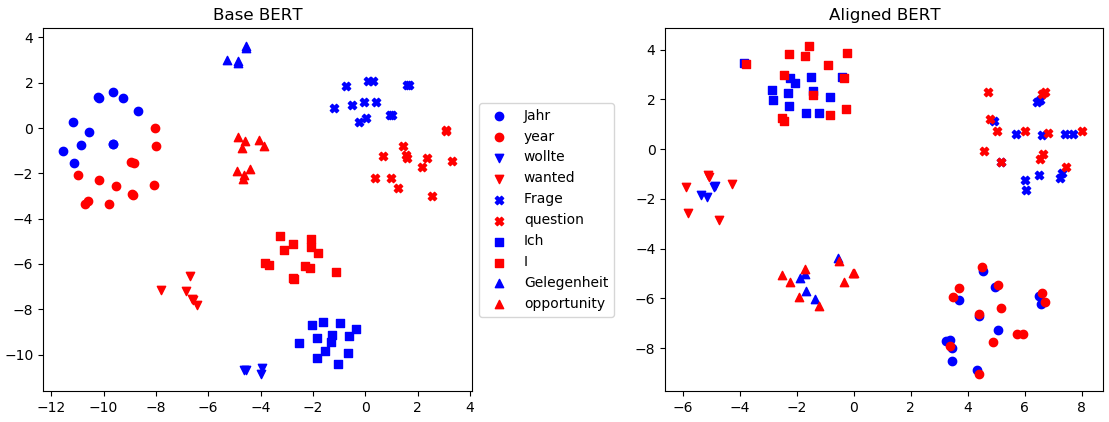}
		\caption{t-SNE~\citep{maaten_tsne} visualization of the embedding space of multilingual BERT for English-German word pairs (left: pre-alignment, right: post-alignment). Each point is a different instance of the word in the Europarl corpus. This figure suggests that BERT begins already somewhat aligned out-of-the-box but becomes much more aligned after our proposed procedure.}
		\label{fig:tsne}
	\end{figure}
	
	Embedding alignment was originally studied for word vectors with the goal of enabling cross-lingual transfer, where the embeddings for two languages are in alignment if word translations, e.g.\ \textit{cat} and \textit{Katze}, have similar representations~\citep{mikolov_alignment_2013,smith_offline_2017}. Recently, large pre-trained models have largely subsumed word vectors based on their accuracy on downstream tasks, partly due to the fact that their word representations are context-dependent, allowing them to more richly capture the meaning of a word~\citep{peters_deep_2018,howard_2018_ulmfit,radford2018improving,devlin_bert:2018}. Therefore, with the same goal of cross-lingual transfer but for these more complex models, we might consider contextual embedding alignment, where we observe whether word pairs within parallel sentences, e.g.\ \textit{cat} in \textit{``The cat sits''} and \textit{Katze} in \textit{``Die Katze sitzt,''} have similar representations.
	
	One model relevant to these questions is multilingual BERT, a version of BERT pre-trained on 104 languages that achieves remarkable transfer on downstream tasks. For example, after the model is fine-tuned on the English MultiNLI training set, it achieves 74.3\% accuracy on the test set in Spanish, which is only 7.1\% lower than the English accuracy~\citep{devlin_bert:2018,conneau-etal-2018-xnli}. Furthermore, while the model transfers better to languages similar to English, it still achieves reasonable accuracies even on languages with different scripts. 
	
	However, given the way that multilingual BERT was pre-trained, it is unclear why we should expect such high zero-shot performance. Compared to monolingual BERT which exhibits no zero-shot transfer, multilingual BERT differs only in that (1) during pre-training (i.e.\ masked word prediction), each batch contains sentences from all of the languages, and (2) it uses a single shared vocabulary, formed by WordPiece on the concatenated monolingual corpora~\citep{devlin_github_2019}. Therefore, we might wonder: (1) How can we better understand BERT's multilingualism? (2) Can we further improve BERT's cross-lingual transfer?
	
	In this paper, we show that contextual embedding alignment is a useful concept for addressing these questions. First, we propose a contextual version of word retrieval to evaluate the degree of alignment, where a model is presented with two parallel corpora, and given a word within a sentence in one corpus, it must find the correct word and sentence in the other. Using this metric of alignment, we show that multilingual BERT achieves zero-shot transfer because its embeddings are partially aligned, as depicted in Figure~\ref{fig:tsne}, with the degree of alignment predicting the degree of downstream transfer. 
	
	Next, using between 10K and 250K sentences per language from the Europarl corpus as parallel data~\citep{koehn-2005-europarl}, we propose a fine-tuning-based alignment procedure and show that it significantly improves BERT as a multilingual model. Specifically, on zero-shot XNLI, where the model is trained on English MultiNLI and tested on other languages~\citep{conneau-etal-2018-xnli}, the aligned model improves accuracies by 2.78\% on average over the base model, and it remarkably matches translate-train models for Bulgarian and Greek, which approximate the fully-supervised setting.
	
	To put our results in the context of past work, we also use word retrieval to compare our fine-tuning procedure to two alternatives: (1) fastText augmented with sentence and aligned using rotations~\citep{bojanowski2017enriching,ruckle_etal_sentemb_2018,artetxe-etal-2018-robust}, and (2) BERT aligned using rotations~\citep{aldarmaki-diab-2019-context,schuster-etal-2019-cross,wang-etal-2019-cross}. We find that when there are multiple occurences per word, fine-tuned BERT outperforms fastText, which outperforms rotation-aligned BERT. This result supports the intuition that contextual alignment is more difficult than its non-contextual counterpart, given that a rotation, at least when applied naively, is no longer sufficient to produce strong alignments. In addition, when there is only one occurrence per word, fine-tuned BERT matches the performance of fastText. Given that context disambiguation is no longer necessary, this result suggests that our fine-tuning procedure is able to align BERT at the type level to a degree that matches non-contextual approaches.
	
	Finally, we use the contextual word retrieval task to conduct finer-grained analysis of multilingual BERT, with the goal of better understanding its strengths and shortcomings. Specifically, we find that base BERT has trouble aligning open-class compared to closed-class parts-of-speech, as well as word pairs that have large differences in usage frequency, suggesting insight into the pre-training procedure that we explore in Section~\ref{section:analysis}. Together, these experiments support contextual alignment as an important task that provides useful insight into large multilingual pre-trained models.
	
	\section{Related Work} 
	\textbf{Word vector alignment.} There has been a long line of works that learn aligned word vectors from varying levels of supervision~\citep{Ruder:2019:SCW:3363500.3363514}. One popular family of methods starts with word vectors learned independently for each language (using a method like skip-gram with negative sampling~\citep{Mikolov:2013:DRW:2999792.2999959}), and it learns a mapping from source language vectors to target language vectors with a bilingual dictionary as supervision~\citep{mikolov_alignment_2013,smith_offline_2017,artetxe-etal-2017-learning}. When the mapping is constrained to be an orthogonal linear transformation, the optimal mapping that minimizes distances between word pairs can be solved in closed form~\citep{artetxe-etal-2016-learning,schonemann_procrustes_1996}. Alignment is evaluated using bilingual lexicon induction, so these papers also propose ways to mitigate the hubness problem in nearest neighbors, e.g.\ by using alternate similarity functions like CSLS~\citep{conneau_word_2018}. A recent set of works has also shown that the mapping can be learned with minimal to no supervision by starting with some minimal seed dictionary and alternating between learning the linear map and inducing the dictionary~\citep{artetxe-etal-2018-robust,conneau_word_2018,hoshen-wolf-2018-non,xu-etal-2018-unsupervised-cross,chen-cardie-2018-unsupervised}.
	
	\textbf{Incorporating context into alignment.} One key challenge in making alignment context aware is that the embeddings are now different across multiple occurrences of the same word. Past papers have handled this issue by removing context and aligning the ``average sense'' of a word. In one such study, \citet{schuster-etal-2019-cross} learn a rotation to align contextual ELMo embeddings~\citep{peters_deep_2018} with the goal of improving zero-shot multilingual dependency parsing, and they handle context by taking the average embedding for a word in all of its contexts. In another paper, \citet{aldarmaki-diab-2019-context} learn a rotation on sentence vectors, produced by taking the average word vector over the sentence, and they show that the resulting alignment also works well for word-level tasks. In a contemporaneous work, \citet{wang-etal-2019-cross} align not only the word but also the context by learning a linear transformation using word-aligned parallel data to align multilingual BERT, with the goal of improving zero-shot dependency parsing numbers. In this paper, we similarly align not only the word but also the context, and we also depart from these past works by using more expressive alignment methods than rotation.
	
	\textbf{Incorporating parallel texts into pre-training.} Instead of performing alignment post-hoc, another line of works proposes contextual pre-training procedures that are more cross-lingually-aware. \citet{wieting-etal-2019-simple} pre-train sentence embeddings using parallel texts by maximizing similarity between sentence pairs while minimizing similarity with negative examples. \citet{lample_xlm_2019} propose a cross-lingual pre-training objective that incorporates parallel data in addition to monolingual corpora, leading to improved downstream cross-lingual transfer. In contrast, our method uses less parallel data and aligns existing pre-trained models rather than requiring pre-training from scratch. 
	
	\textbf{Analyzing multilingual BERT.} \citet{pires-etal-2019-multilingual} present a series of probing experiments to better understand multilingual BERT, and they find that transfer is possible even between dissimilar languages, but that it works better between languages that are typologically similar. They conclude that BERT is remarkably multilingual but falls short for certain language pairs.
	
	\section{Methods}
	\subsection{Multilingual Pre-Training}
	We first briefly describe multilingual BERT~\citep{devlin_bert:2018}. Like monolingual BERT, multilingual BERT is pre-trained on sentences from Wikipedia to perform two tasks: masked word prediction, where it must predict words that are masked within a sentence, and next sentence prediction, where it must predict whether the second sentence follows the first one. The model is trained on 104 languages, with each batch containing training sentences from each language, and it uses a shared vocabulary formed by WordPiece on the 104 Wikipedias concatenated~\citep{wu-etal-2016-wordpiece}.
	\subsection{Defining and Evaluating Contextual Alignment}
	In the following sections, we describe how to define, evaluate, and improve contextual alignment. Given two languages, a model is in \textit{contextual alignment} if it has similar representations for word pairs within parallel sentences. More precisely, suppose we have $N$ parallel sentences $C = \{(\mathbf{s}^1, \mathbf{t}^1),...,(\mathbf{s}^N, \mathbf{t}^N)\}$, where $(\mathbf{s}, \mathbf{t})$ is a source-target sentence pair. Also, let each sentence pair $(\mathbf{s}, \mathbf{t})$ have word pairs, denoted $a(\mathbf{s}, \mathbf{t}) = \{(i_1, j_1),...,(i_m, j_m)\}$, containing position tuples $(i, j)$ such that the words $\mathbf{s}_i$ and $\mathbf{t}_j$ are translations of each other.\footnote{These pairs are called word alignments in the machine translation community, but we use the term ``word pairs'' to avoid confusion with embedding alignment. Also, because BERT operates on subwords while the corpus is aligned at the word level, we keep only the BERT vector for the last subword of each word.} We will use $f$ to represent a pre-trained model such that $f(i, \mathbf{s})$ is the contextual embedding for the $i$th word in $\mathbf{s}$. 
	
	As an example, we might have the following sentence pair:
	\begin{gather*}
	\mathbf{s} = \{\textit{$\stackrel{0}{I\text{\vphantom{l}}}$ $\stackrel{1}{ate\text{\vphantom{l}}}$ $\stackrel{2}{the}$ $\stackrel{3}{apple}$ $\stackrel{4}{.\text{\vphantom{l}}}$}\} \quad \mathbf{t} = \{\textit{$\stackrel{0}{Ich}$ $\stackrel{1}{habe}$ $\stackrel{2}{den}$ $\stackrel{3}{Apfel}$ $\stackrel{4}{gegessen\text{\vphantom{l}}}$ $\stackrel{5}{.\text{\vphantom{l}}}$}\}\\
	a(\mathbf{s}, \mathbf{t}) = \{(0, 0), (1,4), (2,2), (3,3), (4, 5)\}
	\end{gather*}
	
	Then, using the parallel corpus $C$, we can measure the contextual alignment of the model $f$ using its accuracy in \textit{contextual word retrieval}. In this task, the model is presented with two parallel corpora, and given a word within a sentence in one corpus, it must find the correct word and sentence in the other. Specifically, we can define a nearest neighbor retrieval function 
	\begin{equation*}
	\neighbor(i, \mathbf{s}; f, C) = \argmax_{\mathbf{t} \in C,\ 0 \leq j \leq \text{len}(\mathbf{t})} \simfunc(f(i, \mathbf{s}), f(j, \mathbf{t})),
	\end{equation*}
	where $i$ and $j$ denote the position within a sentence and $\simfunc$ is a similarity function. The accuracy is then given by the percentage of exact matches over the entire corpus, or
	\begin{equation*}
	A(f; C) = \frac{1}{N} \sum_{(\mathbf{s}, \mathbf{t}) \in C} \sum_{(i, j) \in a(\mathbf{s}, \mathbf{t})} \mathbb{I}(\neighbor(i, \mathbf{s}; f, C) = (j, \mathbf{t})),
	\end{equation*}
	where $\mathbb{I}$ represents the indicator function. We can perform the same procedure in the other direction, where we pull target words given source words, so we report the average between the two directions. As our similarity function, we use CSLS, a modified version of cosine similarity that mitigates the hubness problem, with neighborhood size $10$~\citep{conneau_word_2018}. One additional point is that this procedure can be made more or less contextual based on the corpus: a corpus with more occurrences for each word type requires better representations of context. Therefore, we also test non-contextual word retrieval by removing all but the first occurrence of each word type. 
	
	Given parallel data, these word pairs can be procured in an unsupervised fashion using standard techniques developed by the machine translation community~\citep{Brown:1993:MSM:972470.972474}. While these methods can be noisy, by running the algorithm in both the source-target and target-source directions and only keeping word pairs in their intersection, we can trade-off coverage for accuracy, producing a reasonably high-precision dataset~\citep{Och:2003:SCV:778822.778824}.
	
	\subsection{Aligning Pre-Trained Contextual Embeddings}
	To improve the alignment of the model $f$ with respect to the corpus $C$, we can encapsulate alignment in the loss function
	\begin{equation*}
	L(f; C) = -\sum_{(\mathbf{s}, \mathbf{t}) \in C} \sum_{(i, j) \in a(\mathbf{s}, \mathbf{t})} \simfunc(f(i, \mathbf{s}), f(j, \mathbf{t})),
	\end{equation*}
	where we sum the similarities between word pairs. Because the CSLS metric is not easily optimized, we instead use the squared error loss, or $\simfunc(f(i, \mathbf{s}), f(j, \mathbf{t})) = -||f(i, \mathbf{s}) - f(j, \mathbf{t})||_2^2$. 
	
	However, note that this loss function does not account for the informativity of $f$; for example, it is zero if $f$ is constant. Therefore, at a high level, we would like to minimize $L(f; C)$ while maintaining some aspect of $f$ that makes it useful, e.g.\ its high accuracy when fine-tuned on downstream tasks. Letting $f_0$ denote the initial pre-trained model before alignment, we achieve this goal by defining a regularization term
	\begin{equation*}
	R(f; C) = \sum_{\mathbf{t} \in C} \sum_{i = 1}^{\text{len}(\mathbf{t})} ||f(j, \mathbf{t}) - f_0(j, \mathbf{t})||_2^2,
	\end{equation*}
	which imposes a penalty if the target language embeddings stray from their initialization. Then, we sample minibatches $B \subset C$ and take gradient steps of the function $L(f; B) + \lambda R(f; B)$ directly on the weights of $f$, which moves the source embeddings toward the target embeddings while preventing the latter from drifting too far. In our experiments, we set $\lambda = 1$.
	
	In the multilingual case, suppose we have $k$ parallel corpora $C^1,...,C^k$, where each corpus has a different source language with the target language as English. Then, we sample equal-sized batches $B^i \subset C^i$ from each corpus and take gradient steps on $\sum_{i = 1}^k L(f; B^i) + \lambda R(f; B^i)$, which moves all of the non-English embeddings toward English.
	
	Note that this alignment method departs from prior work, in which each non-English language is rotated to match the English embedding space through individual learned matrices. Specifically, the most widely used post-hoc alignment method learns a rotation $W$ applied to the source vectors to minimize the distance between parallel word pairs, or
	\begin{equation}\label{eqn:rotate}
	\min_{W} \sum_{(\mathbf{s}, \mathbf{t}) \in C} \sum_{(i, j) \in a(\mathbf{s}, \mathbf{t})} ||W f(i, \mathbf{s}) - f(j, \mathbf{t})||_2^2 \quad s.t. \quad W^\top W = I.
	\end{equation} 
	This problem is known as the Procrustes problem and can be solved in closed form~\citep{schonemann_procrustes_1996}. This approach has the nice property that the vectors are only rotated, preserving distances and therefore the semantic information captured by the embeddings~\citep{artetxe-etal-2016-learning}. However, rotation requires the strong assumption that the embedding spaces are roughly isometric~\citep{sogaard-etal-2018-limitations}, an assumption that may not hold for contextual pre-trained models because they represent more aspects of a word than just its type, i.e.\ context and syntax, which are less likely to be isomorphic between languages. Given that past work has also found independent alignment per language pair to be inferior to joint training~\citep{heyman-etal-2019-learning}, another advantage of our method is that the alignment for all languages is done simultaneously.
	
	As our dataset, we use the Europarl corpora for English paired with Bulgarian, German, Greek, Spanish, and French, the languages represented in both Europarl and XNLI~\citep{koehn-2005-europarl}. After tokenization~\citep{koehn-etal-2007-moses}, we produce word pairs using fastAlign and keep the one-to-one pairs in the intersection~\citep{dyer-etal-2013-simple}. We use the most recent 1024 sentences as the test set, the previous 1024 sentences as the development set, and the following 250K sentences as the training set. Furthermore, we modify the test set accuracy calculation to only include word pairs not seen in the training set. We also remove any exact matches, e.g.\ punctuation and numbers, because BERT is already aligned for these pairs due to its shared vocabulary. Given that parallel data may be limited for low-resource language pairs, we also report numbers for 10K and 50K parallel sentences.
	
	\subsection{Sentence-Augmented Non-Contextual Baseline}
	Given that there has been a long line of work on word vector alignment~\citep[][\textit{inter alia}]{artetxe-etal-2018-robust,conneau_word_2018,smith_offline_2017}, we also compare BERT to a sentence-augmented fastText baseline~\citep{bojanowski2017enriching}. Following \cite{artetxe-etal-2018-robust}, we first normalize, then mean-center, then normalize the word vectors, and we then learn a rotation with the same parallel data as in the contextual case, as described in Equation~\ref{eqn:rotate}. We also strengthen this baseline by including sentence information: specifically, during word retrieval, we concatenate each word vector with a vector representing its sentence. Following~\citet{ruckle_etal_sentemb_2018}, we compute the sentence vector by concatenating the average, maximum, and minimum vector over all of the words in the sentence, a method that was shown to be state-of-the-art for a suite of cross-lingual tasks. We also experimented with other methods, such as first retrieving the sentence and then the word, but found this method resulted in the highest accuracy. As a result, the fastText vectors are $1200$-dimensional, while the BERT vectors are $768$-dimensional.
	
	\subsection{Testing Zero-Shot Transfer}
	The next step is to determine whether better alignment improves cross-lingual transfer. As our downstream task, we use the XNLI dataset, where the English MultiNLI development and test sets are human-translated into multiple languages~\citep{conneau-etal-2018-xnli,williams-etal-2018-broad}. Given a pair of sentences, the task is to predict whether the first sentence implies the second, where there are three labels: entailment, neutral, or contradiction. Starting from either the base or aligned multilingual BERT, we train on English and evaluate on Bulgarian, German, Greek, Spanish, and French, the XNLI languages represented in Europarl.
	
	As our architecture, following~\citet{devlin_bert:2018}, we apply a linear layer followed by softmax on the $\texttt{[CLS]}$ embedding of the sentence pair, producing scores for each of the three labels. The model is trained using cross-entropy loss and selected based on its development set accuracy averaged across all of the languages. As a fully-supervised ceiling, we also compare to models trained and tested on the same language, where for the non-English training data, we use the machine translations of the English MultiNLI training data provided by~\citet{conneau-etal-2018-xnli}. While the quality of the training data is affected by the quality of the MT system, this comparison nevertheless serves as a good approximation for the fully-supervised setting.
	
	\section{Results}
	\subsection{Zero-shot XNLI Transfer}
	\begin{table}[t]
		\begin{center}
			\resizebox{1.0\linewidth}{!}{
			\begin{tabular}{@{}lccccccc@{}}
				\toprule
				& English & Bulgarian & German & Greek & Spanish & French & Average \\
				\midrule
				\midrule
				\multicolumn{3}{@{}l}{Translate-Train} \\
				\midrule
				Base BERT & 81.9 & 73.6 & 75.9 & 71.6 & 77.8 & 76.8 & 76.3 \\
				\midrule
				\midrule
				\multicolumn{3}{@{}l}{Zero-Shot\footnote{Note that the zero-shot Base BERT numbers are slightly different from those reported in \citet{devlin_github_2019} because we select a single model using the average accuracy across the six languages. This selection method also accounts for the varying English accuracies across the zero-shot methods.}} \\
				\midrule
				Base BERT & 80.4 & 68.7 & 70.4 & 67.0 & 74.5 & 73.4 & 72.4 \\
				Sentence-aligned BERT (rotation) & \textbf{81.1} & 68.9 & 71.2 & 66.7 & 74.9 & 73.5 & 72.7 \\
				Word-aligned BERT (rotation) & 78.8 & 69.0 & 71.3 & 67.1 & 74.3 & 73.0 & 72.2 \\
				Word-aligned BERT (fine-tuned) & 80.1 & \textbf{73.4} & \textbf{73.1} & \textbf{71.4} & \textbf{75.5} & \textbf{74.5} & \textbf{74.7} \\
				\midrule
				XLM (MLM + TLM) & \underline{85.0} & \underline{77.4} & \underline{77.8} & \underline{76.6} & \underline{78.9} & \underline{78.7} & \underline{79.1} \\
				\bottomrule
			\end{tabular}
			} 
		\end{center}
		\caption{\label{table:xnli} Accuracy on the XNLI test set, where we compare to base BERT~\citep{devlin_bert:2018} and two rotation-based methods, sentence alignment~\citep{aldarmaki-diab-2019-context} and word alignment~\citep{wang-etal-2019-cross}. We also include the current state-of-the-art zero-shot achieved by XLM~\citep{lample_xlm_2019}. Rotation-based methods provide small gains on some languages but not others. On the other hand, after fine-tuning-based alignment, Bulgarian and Greek match the translate-train ceiling, while German, Spanish, and French close roughly one-third of the gap.}
	\end{table}
	
	First, we test whether alignment improves multilingual BERT by applying the models to zero-shot XNLI, as displayed in Table~\ref{table:xnli}. We see that our alignment procedure greatly improves accuracies, with all languages seeing a gain of at least $1\%$. In particular, the Bulgarian and Greek zero-shot numbers are boosted by almost $5\%$ each and match the translate-train numbers, suggesting that the alignment procedure is especially effective for languages that are initially difficult for BERT. We also run alignment for more distant language pairs (Chinese, Arabic, Urdu) and find similar results, which we report in the appendix.
	
	Comparing to rotation-based methods~\citep{aldarmaki-diab-2019-context,wang-etal-2019-cross}, we find that a rotation produces small gains for some languages, namely Bulgarian, German, and Spanish, but is sub-optimal overall, providing evidence that the increased expressivity of our proposed procedure is beneficial for contextual alignment. We explore this comparison more in Section~\ref{section:retrieval}.
	
	\subsection{Alignment with Less Data}
	\begin{table}[t]
	    \begin{center}
			\begin{tabular}{@{}lccccccc@{}}
				\toprule
				Sentences & English & Bulgarian & German & Greek & Spanish & French & Average \\
				\midrule
				\midrule
				None & 80.4 & 68.7 & 70.4 & 67.0 & 74.5 & 73.4 & 72.4 \\
				10K & 79.2 & 71.0 & 71.8 & 67.5 & 75.3 & 74.1 & 73.2 \\
				50K & \textbf{81.1} & 73 & 72.6 & 69.6 & 75 & \textbf{74.5} & 74.3\\
				250K & 80.1 & \textbf{73.4} & \textbf{73.1} & \textbf{71.4} & \textbf{75.5} & \textbf{74.5} & \textbf{74.7} \\
				\bottomrule
			\end{tabular}
		\end{center}
	\caption{\label{table:xnli_data} Zero-shot accuracy on the XNLI test set, where we align BERT with varying amounts of parallel data. The method scales with the amount of data but achieves a large fraction of the gains with 50K sentences per language pair.}
	\end{table}
	Given that our goal is zero-shot transfer, we cannot expect to always have large amounts of parallel data. Therefore, we also characterize the performance of our alignment method with varying amounts of data, as displayed in Table~\ref{table:xnli_data}. We find that it improves transfer with as little as 10K sentences per language, making it a promising approach for low-resource languages.
	
	\section{Analysis}\label{section:analysis}
	\subsection{Word Retrieval}\label{section:retrieval}
	\begin{table}[t]
	\begin{center}
			\begin{tabular}{@{}lccccccc@{}}
				\toprule
				& bg-en & de-en & el-en & es-en & fr-en & Average\\
				\midrule
				\midrule
				Contextual \\
				\midrule
				Aligned fastText + sentence & 44.0 & 46.4 & 42.0 & 48.6 & 44.5 & 45.1 \\
				Base BERT & 19.5 & 26.1 & 13.9 & 32.5 & 28.3 & 24.1\\
				Word-aligned BERT (rotation) & 29.8 & 31.6 & 20.8 & 36.8 & 31.0 & 30.0 \\
				Word-aligned BERT (fine-tuned) & \textbf{50.7} & \textbf{51.3} & \textbf{49.8} & \textbf{51.0} & \textbf{48.6} & \textbf{50.3} \\
				\midrule
				\midrule
				Non-Contextual \\
				\midrule
				Aligned fastText + sentence & 61.3 & \textbf{65.4} & 61.6 & \textbf{71.1} & 64.8 & 64.8 \\
				Base BERT & 29.1 & 37.0 & 22.3 & 46.5 & 41.8 & 35.3 \\ 
				Word-aligned BERT (rotation) & 39.6 & 43.6 & 32.4 & 51.4 & 46.1 & 42.6 \\
				Word-aligned BERT (fine-tuned) & \textbf{62.8} & 64.3 & \textbf{67.5} & 68.4 & \textbf{66.3} & \textbf{65.9} \\
				\bottomrule
				\bottomrule
			\end{tabular}
	\end{center}
	\caption{\label{table:contextual-word-retrieval} Word retrieval accuracy for the aligned sentence-augmented fastText baseline and BERT pre- and post-alignment. Across languages, base BERT has variable accuracy while fine-tuning-aligned BERT is consistently effective. Fine-tuned BERT also matches fastText in a version of the task where context is not necessary, suggesting that our method matches the type-level alignment of fastText while also aligning context.}
    \end{table}
    
	In the following sections, we present word retrieval results to both compare our method to past work and better understand the strengths and weaknesses of multilingual BERT. Table~\ref{table:contextual-word-retrieval} displays the word retrieval accuracies for the aligned sentence-augmented fastText baseline and BERT pre- and post-alignment. First, we find that in contextual retrieval, fine-tuned BERT outperforms fastText, which outperforms rotation-aligned BERT. This result supports the intuition that aligning large pre-trained models is more difficult than aligning word vectors, given that a rotation, at least when applied naively, produces sub-par alignments. In addition, fine-tuned BERT matches the performance of fastText in non-contextual retrieval, suggesting that our alignment procedure overcomes these challenges and achieves type-level alignment that matches non-contextual approaches. In the appendix, we also provide examples of aligned BERT disambiguating between different meanings of a word, giving qualitative evidence of the benefit of context alignment.
	
	We also find that before alignment, BERT's performance varies greatly between languages, while after alignment it is consistently effective. In particular, Bulgarian and Greek initially have very low accuracies. This phenomenon is also reflected in the XNLI numbers (Table~\ref{table:xnli}), where Bulgarian and Greek receive the largest boosts from alignment. Examining the connection between alignment and zero-shot more closely, we find that the word retrieval accuracies are highly correlated with downstream zero-shot performance (Figure~\ref{fig:plot}), supporting our evaluation measure as predictive of cross-lingual transfer.
	
	The language discrepancies are also consistent with a hypothesis by \citet{pires-etal-2019-multilingual} to explain BERT's multilingualism. They posit that due to the shared vocabulary, shared words between languages, e.g.\ numbers and names, are forced to have the same representation. Then, due to the masked word prediction task, other words that co-occur with these shared words also receive similar representations. If this hypothesis is true, then languages with higher lexical overlap with English are likely to experience higher transfer. As an extreme form of this phenomenon, Bulgarian and Greek have completely different scripts and should experience worse transfer than the common-script languages, an intuition that is confirmed by the word retrieval and XNLI accuracies. The fact that all languages are equally aligned with English post-alignment suggests that the pre-training procedure is suboptimal for these languages.
	
	\subsection{Word Retrieval Part-of-Speech Analysis}
	\begin{table}[t]
		\begin{center}
			\begin{tabular}{@{}lccccc|c@{}}
				\toprule
				Lexical Overlap &  Numeral &  Punctuation &  Proper Noun & & & Average \\
				\midrule
				Base BERT & 0.90 & 0.88 & 0.80 && & 0.86 \\
				Aligned BERT & 0.97 & 0.96 & 0.95 & & & 0.96 \\
				\midrule
				\midrule
				Closed-Class &  Determiner &  Preposition &  Conjunction &  Pronoun &  Auxiliary & Average\\
				\midrule
				Base BERT & 0.76 & 0.72 & 0.71 & 0.70 & 0.61 & 0.70 \\
				Aligned BERT & 0.91 & 0.86 & 0.89 & 0.89 & 0.84 & 0.88 \\
				\midrule
				\midrule
				Open-Class &  Noun &  Adverb &  Adjective &  Verb & & Average\\
				\midrule
				Base BERT & 0.61 & 0.57 & 0.50 & 0.49  & & 0.54 \\
				Aligned BERT & 0.90 & 0.88 & 0.90 & 0.89  & & 0.89 \\
				\bottomrule
			\end{tabular}
		\end{center}
		\caption{\label{table:part-of-speech} Accuracy by part-of-speech tag for non-contextual word retrieval. To achieve better word type coverage, we do not remove word pairs seen in the training set. The tags are grouped into lexically overlapping, closed-class, and open-class groups. The ``Particle,'' ``Symbol,'' ``Interjection,'' and ``Other'' tags are omitted. }
	\end{table}
	
	Next, to gain insight into the multilingual pre-training procedure, we analyze the accuracy broken down by part-of-speech using the Universal Part-of-Speech Tagset~\citep{petrov-etal-2012-universal}, annotated using polyglot~\citep{polyglot:2013:ACL-CoNLL} for Bulgarian and spaCy~\citep{spacy2} for the other languages, as displayed in Table~\ref{table:part-of-speech}. Unsurprisingly, multilingual BERT has high alignment out-of-the-box for groups with high lexical overlap, e.g.\ numerals, punctuation, and proper nouns, due to its shared vocabulary. We further divide the remaining tags into closed-class and open-class, where closed-class parts-of-speech correspond to fixed sets of words serving grammatical functions (e.g.\ determiner, preposition, conjunction, pronoun, and auxiliary), while open-class parts-of-speech correspond to lexical words (e.g.\ noun, adverb, adjective, verb). Interestingly, we see that base BERT has consistently lower accuracy for closed-class versus open-class categories ($0.70$ vs $0.54$), but that this discrepancy disappears after alignment ($0.88$ vs $0.89$).
	
	\subsection{Usage Hypothesis for Alignment}
	\begin{figure}
    \begin{floatrow}
    \ffigbox[0.9\linewidth]{%
      \resizebox{\linewidth}{!}{
				\begin{tikzpicture}
				\begin{axis}[legend pos=outer north east,
				xlabel={Contextual word retrieval accuracy},
				ylabel={XNLI zero-shot accuracy}]
				\addplot [only marks, mark = *] table {\datatable};
				\addplot [thick, red] table[
				y={create col/linear regression={y=Y}}
				] 
				{\datatable};
				\addplot[mark=none] coordinates {(19.5, 68.7)} node[pin=270:{bg}]{} ;
				\addplot[mark=none] coordinates {(26.1, 70.4)} node[pin=270:{de}]{} ;
				\addplot[mark=none] coordinates {(13.9, 67.0)} node[pin=90:{el}]{} ;
				\addplot[mark=none] coordinates {(32.5, 74.5)} node[pin=270:{es}]{} ;
				\addplot[mark=none] coordinates {(28.3, 73.4)} node[pin=90:{fr}]{} ;
				\end{axis}
				\end{tikzpicture}}
    }{%
      \caption{XNLI zero-shot versus word retrieval accuracy for base BERT, where each point is a language paired with English. This plot suggests that alignment correlates well with cross-lingual transfer.}
	    \label{fig:plot}
    }
    \ffigbox[1.1\linewidth]{%
			\includegraphics[width=\linewidth]{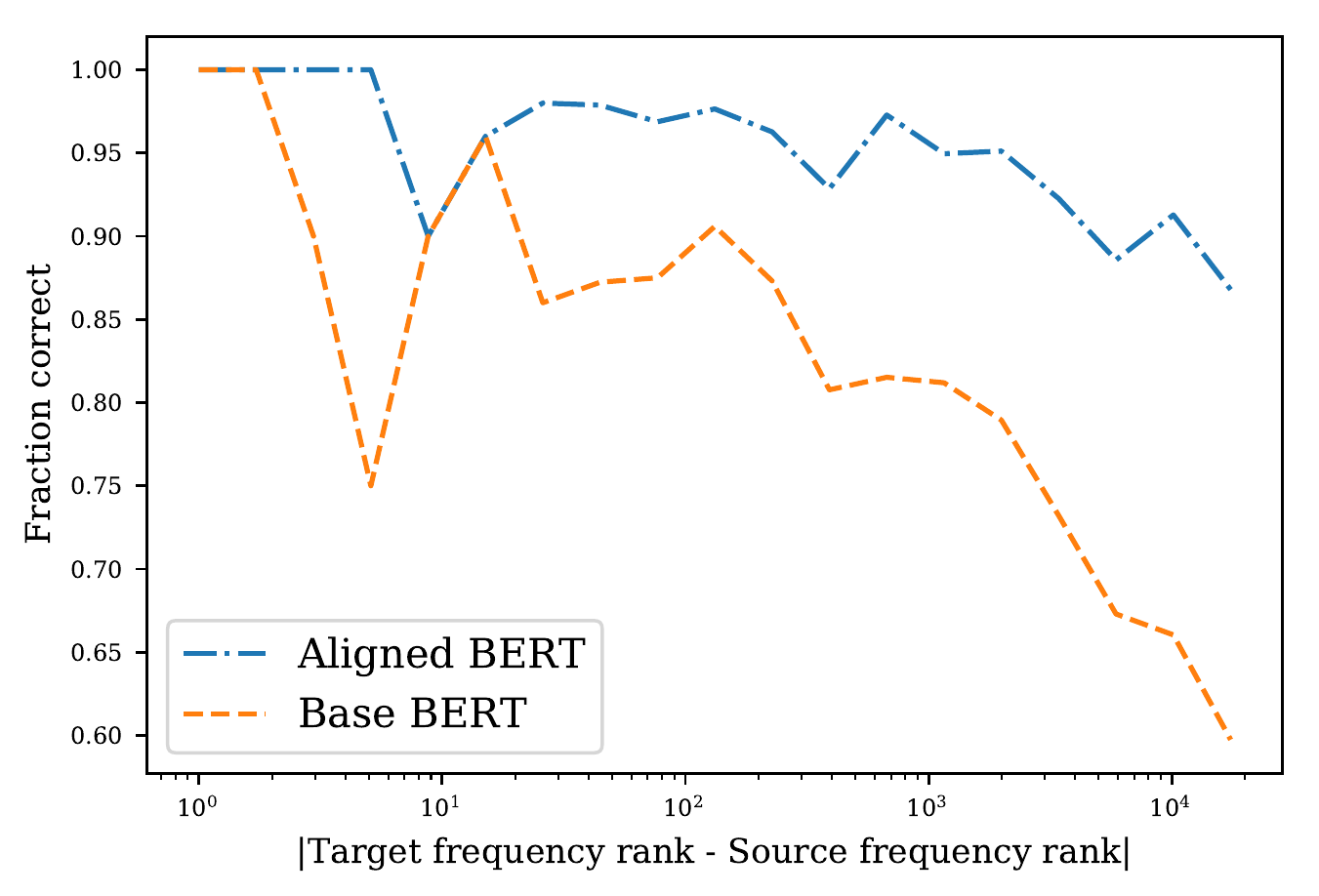}
			
    }{%
    \caption{\label{fig:frequency}Contextual word retrieval accuracy plotted against difference in frequency rank between source and target. The accuracy of base BERT plummets for larger differences, suggesting that its alignment depends on word pairs having similar usage statistics.}
    }
    \end{floatrow}
    \end{figure}
	From this closed-class vs open-class difference, we hypothesize that BERT's alignment of a particular word pair is influenced by the similarity of their usage statistics. Specifically, given that BERT is trained through masked word prediction, its embeddings are in large part determined by the co-occurrences between words. Therefore, two words that are used in similar contexts should be better aligned. This hypothesis provides an explanation of the closed-class vs open-class difference: closed-class words are typically grammatical, so they are used in similar ways across typologically similar languages. Furthermore, these words cannot be substituted for one another due to their grammatical function. Therefore, their usage statistics are a strong signature that can be used for alignment. On the other hand, open-class words can be substituted for one another: for example, in most sentences, the noun tokens could be replaced by a wide range of semantically dissimilar nouns with the sentence remaining syntactically well-formed. By this effect, many nouns have similar co-occurrences, making them difficult to align through masked word prediction alone. 
	
	To further test this hypothesis, we plot the word retrieval accuracy versus the difference between the frequency rank of the target and source word, where this difference measures discrepancies in usage, as depicted in Figure~\ref{fig:frequency}. We see that accuracy drops off significantly as the source-target difference increases, supporting our hypothesis. Furthermore, this shortcoming is remedied by alignment, revealing another systematic deficiency of multilingual pre-training.
	
	
	
	
	
	
	
	\section{Conclusion}
	Given that the degree of alignment is causally predictive of downstream cross-lingual transfer, contextual alignment proves to be a useful concept for understanding and improving multilingual pre-trained models. Given small amounts of parallel data, our alignment procedure improves multilingual BERT and corrects many of its systematic deficiencies. Contextual word retrieval also provides useful new insights into the pre-training procedure, opening up new avenues for analysis.
	

	\bibliography{iclr2020_conference}

\begin{thebibliography}{41}
\providecommand{\natexlab}[1]{#1}
\providecommand{\url}[1]{\texttt{#1}}
\expandafter\ifx\csname urlstyle\endcsname\relax
  \providecommand{\doi}[1]{doi: #1}\else
  \providecommand{\doi}{doi: \begingroup \urlstyle{rm}\Url}\fi

\bibitem[Al-Rfou et~al.(2013)Al-Rfou, Perozzi, and
  Skiena]{polyglot:2013:ACL-CoNLL}
Rami Al-Rfou, Bryan Perozzi, and Steven Skiena.
\newblock Polyglot: Distributed word representations for multilingual nlp.
\newblock In \emph{Proceedings of the Seventeenth Conference on Computational
  Natural Language Learning}, pp.\  183--192, Sofia, Bulgaria, August 2013.
  Association for Computational Linguistics.
\newblock URL \url{http://www.aclweb.org/anthology/W13-3520}.

\bibitem[Aldarmaki \& Diab(2019)Aldarmaki and
  Diab]{aldarmaki-diab-2019-context}
Hanan Aldarmaki and Mona Diab.
\newblock Context-aware cross-lingual mapping.
\newblock In \emph{Proceedings of the 2019 Conference of the North {A}merican
  Chapter of the Association for Computational Linguistics: Human Language
  Technologies, Volume 1 (Long and Short Papers)}, pp.\  3906--3911,
  Minneapolis, Minnesota, June 2019. Association for Computational Linguistics.
\newblock \doi{10.18653/v1/N19-1391}.
\newblock URL \url{https://www.aclweb.org/anthology/N19-1391}.

\bibitem[Artetxe et~al.(2016)Artetxe, Labaka, and
  Agirre]{artetxe-etal-2016-learning}
Mikel Artetxe, Gorka Labaka, and Eneko Agirre.
\newblock Learning principled bilingual mappings of word embeddings while
  preserving monolingual invariance.
\newblock In \emph{Proceedings of the 2016 Conference on Empirical Methods in
  Natural Language Processing}, pp.\  2289--2294, Austin, Texas, November 2016.
  Association for Computational Linguistics.
\newblock \doi{10.18653/v1/D16-1250}.
\newblock URL \url{https://www.aclweb.org/anthology/D16-1250}.

\bibitem[Artetxe et~al.(2017)Artetxe, Labaka, and
  Agirre]{artetxe-etal-2017-learning}
Mikel Artetxe, Gorka Labaka, and Eneko Agirre.
\newblock Learning bilingual word embeddings with (almost) no bilingual data.
\newblock In \emph{Proceedings of the 55th Annual Meeting of the Association
  for Computational Linguistics (Volume 1: Long Papers)}, pp.\  451--462,
  Vancouver, Canada, July 2017. Association for Computational Linguistics.
\newblock \doi{10.18653/v1/P17-1042}.
\newblock URL \url{https://www.aclweb.org/anthology/P17-1042}.

\bibitem[Artetxe et~al.(2018)Artetxe, Labaka, and
  Agirre]{artetxe-etal-2018-robust}
Mikel Artetxe, Gorka Labaka, and Eneko Agirre.
\newblock A robust self-learning method for fully unsupervised cross-lingual
  mappings of word embeddings.
\newblock In \emph{Proceedings of the 56th Annual Meeting of the Association
  for Computational Linguistics (Volume 1: Long Papers)}, pp.\  789--798,
  Melbourne, Australia, July 2018. Association for Computational Linguistics.
\newblock URL \url{https://www.aclweb.org/anthology/P18-1073}.

\bibitem[Bojanowski et~al.(2017)Bojanowski, Grave, Joulin, and
  Mikolov]{bojanowski2017enriching}
Piotr Bojanowski, Edouard Grave, Armand Joulin, and Tomas Mikolov.
\newblock Enriching word vectors with subword information.
\newblock \emph{Transactions of the Association for Computational Linguistics},
  5:\penalty0 135--146, 2017.
\newblock \doi{10.1162/tacl_a_00051}.
\newblock URL \url{https://www.aclweb.org/anthology/Q17-1010}.

\bibitem[Brown et~al.(1993)Brown, Pietra, Pietra, and
  Mercer]{Brown:1993:MSM:972470.972474}
Peter~F. Brown, Vincent J.~Della Pietra, Stephen A.~Della Pietra, and Robert~L.
  Mercer.
\newblock The mathematics of statistical machine translation: Parameter
  estimation.
\newblock \emph{Comput. Linguist.}, 19\penalty0 (2):\penalty0 263--311, June
  1993.
\newblock ISSN 0891-2017.
\newblock URL \url{http://dl.acm.org/citation.cfm?id=972470.972474}.

\bibitem[Chen \& Cardie(2018)Chen and Cardie]{chen-cardie-2018-unsupervised}
Xilun Chen and Claire Cardie.
\newblock Unsupervised multilingual word embeddings.
\newblock In \emph{Proceedings of the 2018 Conference on Empirical Methods in
  Natural Language Processing}, pp.\  261--270, Brussels, Belgium,
  October-November 2018. Association for Computational Linguistics.
\newblock \doi{10.18653/v1/D18-1024}.
\newblock URL \url{https://www.aclweb.org/anthology/D18-1024}.

\bibitem[Conneau et~al.(2018{\natexlab{a}})Conneau, Lample, Ranzato, Denoyer,
  and J'egou]{conneau_word_2018}
Alexis Conneau, Guillaume Lample, Marc'Aurelio Ranzato, Ludovic Denoyer, and
  Herve J'egou.
\newblock Word translation without parallel data.
\newblock In \emph{Proceedings of the 6th International Conference on Learning
  Representations (ICLR 2018)}, 2018{\natexlab{a}}.
\newblock URL \url{https://arxiv.org/pdf/1710.04087.pdf}.

\bibitem[Conneau et~al.(2018{\natexlab{b}})Conneau, Rinott, Lample, Williams,
  Bowman, Schwenk, and Stoyanov]{conneau-etal-2018-xnli}
Alexis Conneau, Ruty Rinott, Guillaume Lample, Adina Williams, Samuel Bowman,
  Holger Schwenk, and Veselin Stoyanov.
\newblock {XNLI}: Evaluating cross-lingual sentence representations.
\newblock In \emph{Proceedings of the 2018 Conference on Empirical Methods in
  Natural Language Processing}, pp.\  2475--2485, Brussels, Belgium,
  October-November 2018{\natexlab{b}}. Association for Computational
  Linguistics.
\newblock \doi{10.18653/v1/D18-1269}.
\newblock URL \url{https://www.aclweb.org/anthology/D18-1269}.

\bibitem[Devlin et~al.(2018)Devlin, Chang, Lee, and
  Toutanova]{devlin_bert:2018}
Jacob Devlin, Ming-Wei Chang, Kenton Lee, and Kristina Toutanova.
\newblock {BERT}: {Pre}-training of deep bidirectional transformers for
  language understanding.
\newblock \emph{arXiv:1810.04805 [cs.CL]}, October 2018.
\newblock URL \url{http://arxiv.org/abs/1810.04805}.

\bibitem[Devlin et~al.(2019)Devlin, Chang, Lee, and
  Toutanova]{devlin_github_2019}
Jacob Devlin, Ming-Wei Chang, Kenton Lee, and Kristina Toutanova.
\newblock {BERT}: {Pre}-training of deep bidirectional transformers for
  language understanding.
\newblock
  \url{https://github.com/google-research/bert/blob/master/multilingual.md},
  2019.

\bibitem[Dyer et~al.(2013)Dyer, Chahuneau, and Smith]{dyer-etal-2013-simple}
Chris Dyer, Victor Chahuneau, and Noah~A. Smith.
\newblock A simple, fast, and effective reparameterization of {IBM} model 2.
\newblock In \emph{Proceedings of the 2013 Conference of the North {A}merican
  Chapter of the Association for Computational Linguistics: Human Language
  Technologies}, pp.\  644--648, Atlanta, Georgia, June 2013. Association for
  Computational Linguistics.
\newblock URL \url{https://www.aclweb.org/anthology/N13-1073}.

\bibitem[Eisele \& Chen(2010)Eisele and Chen]{eisele-chen-2010-multiun}
Andreas Eisele and Yu~Chen.
\newblock {M}ulti{UN}: A multilingual corpus from united nation documents.
\newblock In \emph{Proceedings of the Seventh International Conference on
  Language Resources and Evaluation ({LREC}'10)}, Valletta, Malta, May 2010.
  European Language Resources Association (ELRA).
\newblock URL
  \url{http://www.lrec-conf.org/proceedings/lrec2010/pdf/686_Paper.pdf}.

\bibitem[Heyman et~al.(2019)Heyman, Verreet, Vuli{\'c}, and
  Moens]{heyman-etal-2019-learning}
Geert Heyman, Bregt Verreet, Ivan Vuli{\'c}, and Marie-Francine Moens.
\newblock Learning unsupervised multilingual word embeddings with incremental
  multilingual hubs.
\newblock In \emph{Proceedings of the 2019 Conference of the North {A}merican
  Chapter of the Association for Computational Linguistics: Human Language
  Technologies, Volume 1 (Long and Short Papers)}, pp.\  1890--1902,
  Minneapolis, Minnesota, June 2019. Association for Computational Linguistics.
\newblock \doi{10.18653/v1/N19-1188}.
\newblock URL \url{https://www.aclweb.org/anthology/N19-1188}.

\bibitem[Honnibal \& Montani(2017)Honnibal and Montani]{spacy2}
Matthew Honnibal and Ines Montani.
\newblock {spaCy 2}: Natural language understanding with {B}loom embeddings,
  convolutional neural networks and incremental parsing.
\newblock To appear, 2017.

\bibitem[Hoshen \& Wolf(2018)Hoshen and Wolf]{hoshen-wolf-2018-non}
Yedid Hoshen and Lior Wolf.
\newblock Non-adversarial unsupervised word translation.
\newblock In \emph{Proceedings of the 2018 Conference on Empirical Methods in
  Natural Language Processing}, pp.\  469--478, Brussels, Belgium,
  October-November 2018. Association for Computational Linguistics.
\newblock \doi{10.18653/v1/D18-1043}.
\newblock URL \url{https://www.aclweb.org/anthology/D18-1043}.

\bibitem[Howard \& Ruder(2018)Howard and Ruder]{howard_2018_ulmfit}
Jeremy Howard and Sebastian Ruder.
\newblock Universal language model fine-tuning for text classification.
\newblock In \emph{Proceedings of the 56th Annual Meeting of the Association
  for Computational Linguistics (Volume 1: Long Papers)}, pp.\  328--339.
  Association for Computational Linguistics, 2018.
\newblock URL \url{http://aclweb.org/anthology/P18-1031}.

\bibitem[Koehn(2005)]{koehn-2005-europarl}
Philipp Koehn.
\newblock Europarl: A parallel corpus for statistical machine translation.
\newblock In \emph{Conference Proceedings: The Tenth Machine Translation
  Summit}, pp.\  79--86, Phuket, Thailand, 2005. AAMT.

\bibitem[Koehn et~al.(2007)Koehn, Hoang, Birch, Callison-Burch, Federico,
  Bertoldi, Cowan, Shen, Moran, Zens, Dyer, Bojar, Constantin, and
  Herbst]{koehn-etal-2007-moses}
Philipp Koehn, Hieu Hoang, Alexandra Birch, Chris Callison-Burch, Marcello
  Federico, Nicola Bertoldi, Brooke Cowan, Wade Shen, Christine Moran, Richard
  Zens, Chris Dyer, Ond{\v{r}}ej Bojar, Alexandra Constantin, and Evan Herbst.
\newblock {M}oses: Open source toolkit for statistical machine translation.
\newblock In \emph{Proceedings of the 45th Annual Meeting of the Association
  for Computational Linguistics Companion Volume Proceedings of the Demo and
  Poster Sessions}, pp.\  177--180, Prague, Czech Republic, June 2007.
  Association for Computational Linguistics.
\newblock URL \url{https://www.aclweb.org/anthology/P07-2045}.

\bibitem[Lample \& Conneau(2019)Lample and Conneau]{lample_xlm_2019}
Guillame Lample and Alexis Conneau.
\newblock Cross-lingual language model pretraining.
\newblock 2019.
\newblock URL \url{https://arxiv.org/pdf/1901.07291.pdf}.

\bibitem[Maaten \& Hinton(2008)Maaten and Hinton]{maaten_tsne}
Laurens van~der Maaten and Geoffrey Hinton.
\newblock Visualizing data using t-sne.
\newblock \emph{Journal of Machine Learning Research}, 9:\penalty0 2579--2605,
  2008.
\newblock URL \url{http://www.jmlr.org/papers/v9/vandermaaten08a.html}.

\bibitem[Mikolov et~al.(2013{\natexlab{a}})Mikolov, Le, and
  Sutskever]{mikolov_alignment_2013}
Tomas Mikolov, Quoc~V Le, and Ilya Sutskever.
\newblock Exploiting similarities among languages for machine translation.
\newblock 2013{\natexlab{a}}.
\newblock URL \url{https://arxiv.org/pdf/1309.4168.pdf}.

\bibitem[Mikolov et~al.(2013{\natexlab{b}})Mikolov, Sutskever, Chen, Corrado,
  and Dean]{Mikolov:2013:DRW:2999792.2999959}
Tomas Mikolov, Ilya Sutskever, Kai Chen, Greg Corrado, and Jeffrey Dean.
\newblock Distributed representations of words and phrases and their
  compositionality.
\newblock In \emph{Proceedings of the 26th International Conference on Neural
  Information Processing Systems - Volume 2}, NIPS'13, pp.\  3111--3119, USA,
  2013{\natexlab{b}}. Curran Associates Inc.
\newblock URL \url{http://dl.acm.org/citation.cfm?id=2999792.2999959}.

\bibitem[Och \& Ney(2003)Och and Ney]{Och:2003:SCV:778822.778824}
Franz~Josef Och and Hermann Ney.
\newblock A systematic comparison of various statistical alignment models.
\newblock \emph{Comput. Linguist.}, 29\penalty0 (1):\penalty0 19--51, March
  2003.
\newblock ISSN 0891-2017.
\newblock \doi{10.1162/089120103321337421}.
\newblock URL \url{http://dx.doi.org/10.1162/089120103321337421}.

\bibitem[Peters et~al.(2018)Peters, Neumann, Iyyer, Gardner, Clark, Lee, and
  Zettlemoyer]{peters_deep_2018}
Matthew Peters, Mark Neumann, Mohit Iyyer, Matt Gardner, Christopher Clark,
  Kenton Lee, and Luke Zettlemoyer.
\newblock Deep contextualized word representations.
\newblock In \emph{Proceedings of the 2018 Conference of the North {A}merican
  Chapter of the Association for Computational Linguistics: Human Language
  Technologies, Volume 1 (Long Papers)}, pp.\  2227--2237, New Orleans,
  Louisiana, June 2018. Association for Computational Linguistics.
\newblock \doi{10.18653/v1/N18-1202}.
\newblock URL \url{https://www.aclweb.org/anthology/N18-1202}.

\bibitem[Petrov et~al.(2012)Petrov, Das, and
  McDonald]{petrov-etal-2012-universal}
Slav Petrov, Dipanjan Das, and Ryan McDonald.
\newblock A universal part-of-speech tagset.
\newblock In \emph{Proceedings of the Eighth International Conference on
  Language Resources and Evaluation ({LREC}-2012)}, pp.\  2089--2096, Istanbul,
  Turkey, May 2012. European Languages Resources Association (ELRA).
\newblock URL
  \url{http://www.lrec-conf.org/proceedings/lrec2012/pdf/274_Paper.pdf}.

\bibitem[Pires et~al.(2019)Pires, Schlinger, and
  Garrette]{pires-etal-2019-multilingual}
Telmo Pires, Eva Schlinger, and Dan Garrette.
\newblock How multilingual is multilingual {BERT}?
\newblock In \emph{Proceedings of the 57th Annual Meeting of the Association
  for Computational Linguistics}, pp.\  4996--5001, Florence, Italy, July 2019.
  Association for Computational Linguistics.
\newblock URL \url{https://www.aclweb.org/anthology/P19-1493}.

\bibitem[Radford et~al.(2018)Radford, Narasimhan, Salimans, and
  Sutskever]{radford2018improving}
Alec Radford, Karthik Narasimhan, Tim Salimans, and Ilya Sutskever.
\newblock Improving language understanding by generative pre-training.
\newblock 2018.
\newblock URL
  \url{https://s3-us-west-2.amazonaws.com/openai-assets/research-covers/language-unsupervised/language\_understanding\_paper.pdf}.

\bibitem[R{\"{u}}ckl{\'{e}} et~al.(2018)R{\"{u}}ckl{\'{e}}, Eger, Peyrard, and
  Gurevych]{ruckle_etal_sentemb_2018}
Andreas R{\"{u}}ckl{\'{e}}, Steffen Eger, Maxime Peyrard, and Iryna Gurevych.
\newblock Concatenated p-mean word embeddings as universal cross-lingual
  sentence representations.
\newblock \emph{arXiv:1803.01400 [cs.CL]}, 2018.
\newblock URL \url{http://arxiv.org/abs/1803.01400}.

\bibitem[Ruder et~al.(2019)Ruder, Vuli\'{c}, and
  S{\o}gaard]{Ruder:2019:SCW:3363500.3363514}
Sebastian Ruder, Ivan Vuli\'{c}, and Anders S{\o}gaard.
\newblock A survey of cross-lingual word embedding models.
\newblock \emph{J. Artif. Int. Res.}, 65\penalty0 (1):\penalty0 569--630, May
  2019.
\newblock ISSN 1076-9757.
\newblock \doi{10.1613/jair.1.11640}.
\newblock URL \url{https://doi.org/10.1613/jair.1.11640}.

\bibitem[Schonemann(1966)]{schonemann_procrustes_1996}
Peter~H. Schonemann.
\newblock A generalized solution of the orthogonal procrustes problem.
\newblock \emph{Psychometrika}, 31(1):\penalty0 1--10, 1966.

\bibitem[Schuster et~al.(2019)Schuster, Ram, Barzilay, and
  Globerson]{schuster-etal-2019-cross}
Tal Schuster, Ori Ram, Regina Barzilay, and Amir Globerson.
\newblock Cross-lingual alignment of contextual word embeddings, with
  applications to zero-shot dependency parsing.
\newblock In \emph{Proceedings of the 2019 Conference of the North {A}merican
  Chapter of the Association for Computational Linguistics: Human Language
  Technologies, Volume 1 (Long and Short Papers)}, pp.\  1599--1613,
  Minneapolis, Minnesota, June 2019. Association for Computational Linguistics.
\newblock \doi{10.18653/v1/N19-1162}.
\newblock URL \url{https://www.aclweb.org/anthology/N19-1162}.

\bibitem[Smith et~al.(2017)Smith, Turban, Hamblin, and
  Hammerla]{smith_offline_2017}
Samuel~L. Smith, David H.~P. Turban, Steven Hamblin, and Nils~Y. Hammerla.
\newblock Offline bilingual word vectors, orthogonal transformations and the
  inverted softmax.
\newblock In \emph{Proceedings of the 5th International Conference on Learning
  Representations (ICLR 2017)}, 2017.
\newblock URL \url{https://openreview.net/pdf?id=r1Aab85gg}.

\bibitem[S{\o}gaard et~al.(2018)S{\o}gaard, Ruder, and
  Vuli{\'c}]{sogaard-etal-2018-limitations}
Anders S{\o}gaard, Sebastian Ruder, and Ivan Vuli{\'c}.
\newblock On the limitations of unsupervised bilingual dictionary induction.
\newblock In \emph{Proceedings of the 56th Annual Meeting of the Association
  for Computational Linguistics (Volume 1: Long Papers)}, pp.\  778--788,
  Melbourne, Australia, July 2018. Association for Computational Linguistics.
\newblock \doi{10.18653/v1/P18-1072}.
\newblock URL \url{https://www.aclweb.org/anthology/P18-1072}.

\bibitem[Tiedemann(2012)]{tiedemann-2012-parallel}
J{\"o}rg Tiedemann.
\newblock Parallel data, tools and interfaces in {OPUS}.
\newblock In \emph{Proceedings of the Eighth International Conference on
  Language Resources and Evaluation ({LREC}'12)}, pp.\  2214--2218, Istanbul,
  Turkey, May 2012. European Language Resources Association (ELRA).
\newblock URL
  \url{http://www.lrec-conf.org/proceedings/lrec2012/pdf/463_Paper.pdf}.

\bibitem[Wang et~al.(2019)Wang, Che, Guo, Liu, and Liu]{wang-etal-2019-cross}
Yuxuan Wang, Wanxiang Che, Jiang Guo, Yijia Liu, and Ting Liu.
\newblock Cross-lingual {BERT} transformation for zero-shot dependency parsing.
\newblock In \emph{Proceedings of the 2019 Conference on Empirical Methods in
  Natural Language Processing and the 9th International Joint Conference on
  Natural Language Processing (EMNLP-IJCNLP)}, pp.\  5725--5731, Hong Kong,
  China, November 2019. Association for Computational Linguistics.
\newblock \doi{10.18653/v1/D19-1575}.
\newblock URL \url{https://www.aclweb.org/anthology/D19-1575}.

\bibitem[Wieting et~al.(2019)Wieting, Gimpel, Neubig, and
  Berg-Kirkpatrick]{wieting-etal-2019-simple}
John Wieting, Kevin Gimpel, Graham Neubig, and Taylor Berg-Kirkpatrick.
\newblock Simple and effective paraphrastic similarity from parallel
  translations.
\newblock In \emph{Proceedings of the 57th Annual Meeting of the Association
  for Computational Linguistics}, pp.\  4602--4608, Florence, Italy, July 2019.
  Association for Computational Linguistics.
\newblock \doi{10.18653/v1/P19-1453}.
\newblock URL \url{https://www.aclweb.org/anthology/P19-1453}.

\bibitem[Williams et~al.(2018)Williams, Nangia, and
  Bowman]{williams-etal-2018-broad}
Adina Williams, Nikita Nangia, and Samuel Bowman.
\newblock A broad-coverage challenge corpus for sentence understanding through
  inference.
\newblock In \emph{Proceedings of the 2018 Conference of the North {A}merican
  Chapter of the Association for Computational Linguistics: Human Language
  Technologies, Volume 1 (Long Papers)}, pp.\  1112--1122, New Orleans,
  Louisiana, June 2018. Association for Computational Linguistics.
\newblock \doi{10.18653/v1/N18-1101}.
\newblock URL \url{https://www.aclweb.org/anthology/N18-1101}.

\bibitem[Wu et~al.(2016)Wu, Schuster, Chen, Le, Norouzi, Macherey, Krikun, Cao,
  Gao, Macherey, Klingner, Shah, Johnson, Liu, Kaiser, Gouws, Kato, Kudo,
  Kazawa, Stevens, Kurian, Patil, Wang, Young, Smith, Riesa, Rudnick, Vinyals,
  Corrado, Hughes, and Dean]{wu-etal-2016-wordpiece}
Yonghui Wu, Mike Schuster, Zhifeng Chen, Quoc~V. Le, Mohammad Norouzi, Wolfgang
  Macherey, Maxim Krikun, Yuan Cao, Qin Gao, Klaus Macherey, Jeff Klingner,
  Apurva Shah, Melvin Johnson, Xiaobing Liu, Lukasz Kaiser, Stephan Gouws,
  Yoshikiyo Kato, Taku Kudo, Hideto Kazawa, Keith Stevens, George Kurian,
  Nishant Patil, Wei Wang, Cliff Young, Jason Smith, Jason Riesa, Alex Rudnick,
  Oriol Vinyals, Greg Corrado, Macduff Hughes, and Jeffrey Dean.
\newblock Google's neural machine translation system: Bridging the gap between
  human and machine translation.
\newblock \emph{arXiv:1609.08144 [cs.CL]}, 2016.

\bibitem[Xu et~al.(2018)Xu, Yang, Otani, and
  Wu]{xu-etal-2018-unsupervised-cross}
Ruochen Xu, Yiming Yang, Naoki Otani, and Yuexin Wu.
\newblock Unsupervised cross-lingual transfer of word embedding spaces.
\newblock In \emph{Proceedings of the 2018 Conference on Empirical Methods in
  Natural Language Processing}, pp.\  2465--2474, Brussels, Belgium,
  October-November 2018. Association for Computational Linguistics.
\newblock \doi{10.18653/v1/D18-1268}.
\newblock URL \url{https://www.aclweb.org/anthology/D18-1268}.

\end{thebibliography}
	\bibliographystyle{iclr2020_conference}
	
	\appendix
	\section{Appendix}
	\subsection{Optimization Hyperparameters}
	For both alignment and XNLI optimization, we use a learning rate of $5 \times 10^{-5}$ with Adam hyperparameters $\beta = (0.9, 0.98)$, $\epsilon = 10^{-9}$ and linear learning rate warmup for the first $10\%$ of the training data. For alignment, the model is trained for one epoch, with each batch containing $2$ sentence pairs per language. For XNLI, each model is trained for $3$ epochs with $32$ examples per batch, and $10\%$ dropout is applied to the BERT embeddings.
	\subsection{Alignment of Chinese, Arabic, and Urdu}
	\begin{table}[t]
	    \begin{center}
	    \resizebox{\linewidth}{!}{
			\begin{tabular}{@{}lcccccccccc@{}}
				\toprule
				 & English & Bulgarian & German & Greek & Spanish & French & Arabic & Chinese & Urdu & Average \\
				\midrule
				\midrule
				Translate-Train \\
				\midrule
				Base BERT  & 81.9 & 73.6 & 75.9 & 71.6 & 77.8 & 76.8 & 70.7 & 76.6 & 61.6 & 74.1 \\
				\midrule
				\midrule
				Zero-Shot \\
				\midrule
				Base BERT & 80.4 & 68.7 & 70.4 & 67.0 & 74.5 & 73.4 & 65.6 & 70.6 & 60.3 & 70.1 \\
				Aligned BERT (20K sent) & \textbf{80.8} & \textbf{71.6} & \textbf{72.5} & \textbf{68.1} & \textbf{74.7} & \textbf{73.6} & \textbf{66.3} & \textbf{71.5} & \textbf{61.1} & \textbf{71.1}\\
				\bottomrule
			\end{tabular}}
		\end{center}
	\caption{\label{table:more_lang} Zero-shot accuracy on the XNLI test set with more languages, where we use 20K parallel sentences for each language paired with English. This result confirms that the alignment method works for distant languages and a variety of parallel corpora, including Europarl, MultiUN, and Tanzil, which contains sentences from the Quran~\citep{koehn-2005-europarl,eisele-chen-2010-multiun,tiedemann-2012-parallel}.}
	\end{table}
	In Table~\ref{table:more_lang}, we report numbers for additional languages, where we align a single BERT model for all eight languages and then fine-tune on XNLI. We use 20K sentences per language, where we use the MultiUN corpus for Arabic and Chinese~\citep{eisele-chen-2010-multiun}, the Tanzil corpus for Urdu~\citep{tiedemann-2012-parallel}, and the Europarl corpus for the other five languages~\citep{koehn-2005-europarl}. This result confirms that the alignment method works for a variety of languages and corpora. Furthermore, the Tanzil corpus consists of sentences from the Quran, suggesting that the method works even when the parallel corpus and downstream task contain sentences from entirely different domains.
	\subsection{Examples of Context-Aware Retrieval}
	In this section, we qualitatively show that aligned BERT is able to disambiguate between different occurences of a word. 
	
	First, we find two meanings of the word ``like'' occurring in the English-German Europarl test set. Note also that in the second and third example, the two senses of ``like'' occur in the same sentence.
	\begin{itemize}
		\item This empire did not look for colonies far from home or overseas, \textbf{like} most Western European States, but close by. 
		
		Dieses Reich suchte seine Kolonien nicht weit von zu Hause und in Übersee \textbf{wie} die meisten westeurop\"{a}ischen Staaten, sondern in der unmittelbaren Umgebung. 
		
		\item \textbf{Like} other speakers, I would like to support the call for the arms embargo to remain. 
		
		\textbf{Wie} andere Sprecher, so m\"{o}chte auch ich den Aufruf zur Aufrechterhaltung des Waffenembargos unterstützen. 
		
		\item Like other speakers, I would \textbf{like} to support the call for the arms embargo to remain. 
		
		Wie andere Sprecher, so \textbf{m\"{o}chte} auch ich den Aufruf zur Aufrechterhaltung des Waffenembargos unterstützen. 
		
		\item I would also \textbf{like}, although they are absent, to mention the Commission and the Council. 
		
		Ich \textbf{m\"{o}chte} mir sogar erlauben, die Kommission und den Rat zu nennen, auch wenn sie nicht anwesend sind.
	\end{itemize}
	\newpage Multiple meanings of ``order'':
	\begin{itemize}
		\item Moreover, the national political elite had to make a detour in Ambon in \textbf{order} to reach the civil governor's residence by warship.
		
		In Ambon mu{\ss}te die politische Spitze des Landes auch noch einen Umweg machen, \textbf{um} mit einem Kriegsschiff die Residenz des Provinzgouverneurs zu erreichen.
		
		\item Although the European Union has an interest in being surrounded by large, stable regions, the tools it has available in \textbf{order} to achieve this are still very limited.
		
		Der Europ\"{a}ischen Union ist zwar an gro{\ss}en stabilen Regionen in ihrer Umgebung gelegen, aber sie verfügt nach wie vor nur über recht begrenzte Instrumente, \textbf{um} das zu erreichen.
		
		\item We could reasonably expect the new Indonesian government to take action in three fundamental areas: restoring public \textbf{order}, prosecuting and punishing those who have blood on their hands and entering into a political dialogue with the opposition.
		
		Von der neuen indonesischen Regierung darf man mit Fug und Recht drei elementare Ma{\ss}nahmen erwarten: die Wiederherstellung der \"{o}ffentlichen \textbf{Ordnung}, die Verfolgung und Bestrafung derjenigen, an deren H\"{a}nden Blut klebt, und die Aufnahme des politischen Dialogs mit den Gegnern.
		
		\item Firstly, I might mention the fact that the army needs to be reformed, secondly that a stable system of law and \textbf{order} needs to be introduced.
		
		Ich nenne hier an erster Stelle die notwendige Reform der Armee, ferner die Einführung eines stabilen Systems rechtsstaatlicher \textbf{Ordnung}.
		
	\end{itemize}
	
	Multiple meanings of ``support'':
	\begin{itemize}
		\item Financial \textbf{support} is needed to enable poor countries to take part in these court activities.
		
		Arme L\"{a}nder m\"{u}ssen finanziell \textbf{unterst\"{u}tzt} werden, damit auch sie sich an der Arbeit des Gerichtshofs beteiligen k\"{o}nnen.
		
		\item We must help them and ensure that a proper action plan is implemented to \textbf{support} their work.
		
		Es gilt einen wirklichen Aktionsplan auf den Weg zu bringen, um die Arbeit dieser Organisationen zu \textbf{unterst\"{u}tzen}.
		
		\item So I hope that you will all \textbf{support} this resolution condemning the abominable conditions of prisoners and civilians in Djibouti.
		
		Ich hoffe daher, da{\ss} Sie alle diese Entschlie{\ss}ung \textbf{bef\"{u}rworten}, die die entsetzlichen Bedingungen von Inhaftierten und Zivilpersonen in Dschibuti verurteilt.
		
		\item It would be difficult to \textbf{support} a subsidy scheme that channelled most of the aid to the large farms in the best agricultural regions.
		
		Es w\"{a}re auch problematisch, ein Beihilfesystem zu \textbf{bef\"{u}rworten}, das die meisten Beihilfen in die gro{\ss}en Betriebe in den besten landwirtschaftlichen Gebieten lenkt.
	\end{itemize}
	
	Multiple meanings of ``close'':
	\begin{itemize}
		\item This empire did not look for colonies far from home or overseas, like most Western European States, but \textbf{close} by. 
		
		Dieses Reich suchte seine Kolonien nicht weit von zu Hause und in Übersee wie die meisten westeurop\"{a}ischen Staaten, sondern in der unmittelbaren \textbf{Umgebung}. 
		
		\item In addition, if we are to shut down or refuse investment from every company which may have an association with the arms industry, then we would have to \textbf{close} virtually every American and Japanese software company on the island of Ireland with catastrophic consequences. 
		
		Wenn wir zudem jedes Unternehmen, das auf irgendeine Weise mit der Rüstungsindustrie verbunden ist, schlie{\ss}en oder Investitionen dieser Unternehmen unterbinden, dann mü{\ss}ten wir so ziemlich alle amerikanischen und japanischen Softwareunternehmen auf der irischen Insel \textbf{schlie{\ss}en}, was katastrophale Auswirkungen h\"{a}tte. 
		
		\item On the other hand, the deployment of resources left over in the Structural Funds from the programme planning period 1994 to 1999 is hardly worth considering as the available funds have already been allocated to specific measures, in this case in \textbf{close} collaboration with the relevant French authorities. 
		
		Die Verwendung verbliebener Mittel der Strukturfonds aus dem Programmplanungszeitraum 1994 bis 1999 ist dagegen kaum in Erw\"{a}gung zu ziehen, da die verfügbaren Mittel bereits bestimmten Ma{\ss}nahmen zugewiesen sind, und zwar im konkreten Fall im \textbf{engen} Zusammenwirken mit den zust\"{a}ndigen franz\"{o}sischen Beh\"{o}rden. 
		
		\item This is particularly justified given that, as already stated, many Member States have very \textbf{close} relations with Djibouti. 
		
		Zumal, wie erw\"{a}hnt, viele Mitgliedstaaten sehr \textbf{enge} Beziehungen zu Dschibuti unterhalten. 
		
		\item Mr President, it is regrettable that, at the \textbf{close} of the 20th century, a century symbolised so positively by the peaceful women's revolution, there are still countries, such as Kuwait and Afghanistan, where half the population, women that is, is still denied fundamental human rights.
		
		Herr Pr\"{a}sident! Es ist wirklich bedauerlich, da{\ss} es am \textbf{Ende} des 20. Jahrhunderts, eines so positiv von der friedlichen Revolution der Frauen gepr\"{a}gten Jahrhunderts, noch immer L\"{a}nder wie Kuwait und Afghanistan gibt, in denen der H\"{a}lfte der Bev\"{o}lkerung, den Frauen, die elementaren Menschenrechte verweigert werden. 
	\end{itemize}
	
\end{document}